\documentclass[10pt,twocolumn,letterpaper]{article}

\usepackage{iccv}
\usepackage{times}
\usepackage{epsfig}
\usepackage{graphicx}
\usepackage{amsmath}
\usepackage{amssymb}

\usepackage{multirow}
\usepackage{amsthm}
\usepackage{booktabs}

\usepackage{makecell}  
\usepackage{scalerel} 
\usepackage{diagbox} 
\usepackage[ruled,vlined]{algorithm2e}
\usepackage{subfig} 
\usepackage[export]{adjustbox} 
\usepackage{wrapfig} 
\usepackage{placeins}

\usepackage[pagebackref=true,breaklinks=true,letterpaper=true,colorlinks,bookmarks=false]{hyperref}

\iccvfinalcopy 

\def\iccvPaperID{****} 

\ificcvfinal\fi
\begin{document}

\title{Fix Your Features: Stationary and Maximally Discriminative Embeddings using Regular Polytope (Fixed Classifier) Networks \vspace{-0.5cm}}

\author{Federico Pernici, Matteo Bruni, Claudio Baecchi and Alberto Del Bimbo \\
MICC -- Media Integration and Communication Center\\
University of Florence -- Italy\\
{\tt\small \{federico.pernici, matteo.bruni, claudio.baecchi, alberto.delbimbo\}@unifi.it}
\vspace{-0.5cm}
}

\maketitle

\begin{abstract}
Neural networks are widely used as a model for classification in a large variety of tasks. Typically, a learnable transformation (i.e. the classifier) is placed at the end of such models returning a value for each class used for classification. This transformation plays an important role in determining how the generated features change during the learning process.
\\
\indent
In this work we argue that this transformation not only can be fixed (i.e. set as non trainable) with no loss of accuracy, but it can also be used to learn stationary and maximally discriminative embeddings.
\\
\indent
We show that the stationarity of the embedding and its maximal  discriminative representation can be theoretically justified by setting the weights of the fixed classifier to values taken from the coordinate vertices of three regular polytopes available in $\mathbb{R}^d$, namely: the $d$-Simplex, the $d$-Cube and the $d$-Orthoplex. These regular polytopes have the maximal amount of symmetry that can be exploited to generate stationary features angularly centered around their corresponding fixed weights.
\\
\indent
Our approach improves and broadens the concept of a fixed classifier, recently proposed in \cite{hoffer2018fix}, to a larger class of fixed classifier models. 
Experimental results confirm both the theoretical analysis and the generalization capability of the proposed method.   
\end{abstract}

\section{Introduction}
Deep Convolutional Neural Networks (DCNNs) have achieved state-of-the-art performance on several classification and representation  tasks in Computer Vision \cite{Zoph_2018_CVPR, Cao18}. In DCNNs, both representation and classification are jointly learned in a single network. The representation for an input sample is the feature vector $\mathbf{f}$ generated by the penultimate layer, while the last layer (i.e. the classifier) outputs score values according to the inner product as: 
\begin{equation}
z_i = \mathbf{w}_i^\top \cdot \mathbf{f}
\label{eq_logit}
\end{equation}
for each class $i$, where $\mathbf{w}_i$ is the weight vector of the classifier for the class $i$. To evaluate the loss, the scores are further normalized into probabilities via the Softmax function \cite{goodfellow2016deep}.
\begin{figure}[t]
\includegraphics[width=0.99\columnwidth]{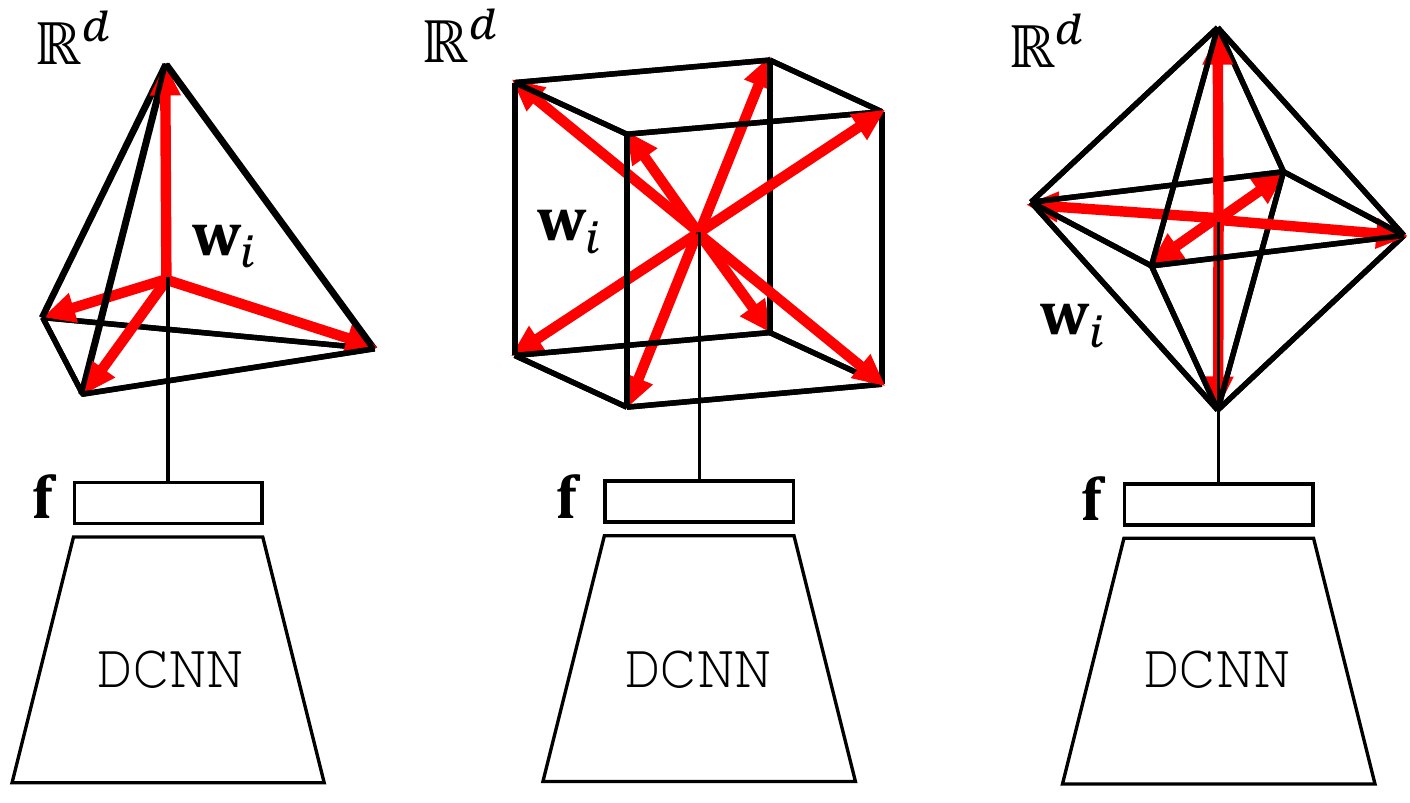}
\caption{ Regular Polytope Networks (RePoNet). 
The fixed classifiers derived from the three regular polytopes available in $\mathbb{R}^d$ with $d \geq 5$ are shown. 
From left: the $d$-Simplex, the $d$-Cube and the $d$-Orthoplex fixed classifier. The trainable parameters $\mathbf{w}_i$ of the classifier are replaced with fixed values taken from the coordinate vertices of a regular polytope (shown in red).
\vspace{-0.3cm}
}
\label{fig_IntroFig}
\end{figure}

Since the values of $z_i$ can be also expressed as $z_i = \mathbf{w}^\top_i \cdot \mathbf{f} = ||\mathbf{w}_i|| \: ||\mathbf{f}|| \cos(\theta)$, where $\theta$ is the angle between $\mathbf{w}_i$ and $\mathbf{f}$, the score for the correct label with respect to the other labels is obtained by optimizing $||\mathbf{w}_i||$, $||\mathbf{f}||$ and $\theta$. 
This simple formulation of the final classifier provides the intuitive explanation of how feature vector directions and weight vector directions align simultaneously with each other at training time so that their average angle is made as small as possible.
\begin{figure*}[htbp]
\hspace{-0.56cm}
\centering
\subfloat[]{\label{fig:a}\includegraphics[height=0.181\linewidth,valign=t]{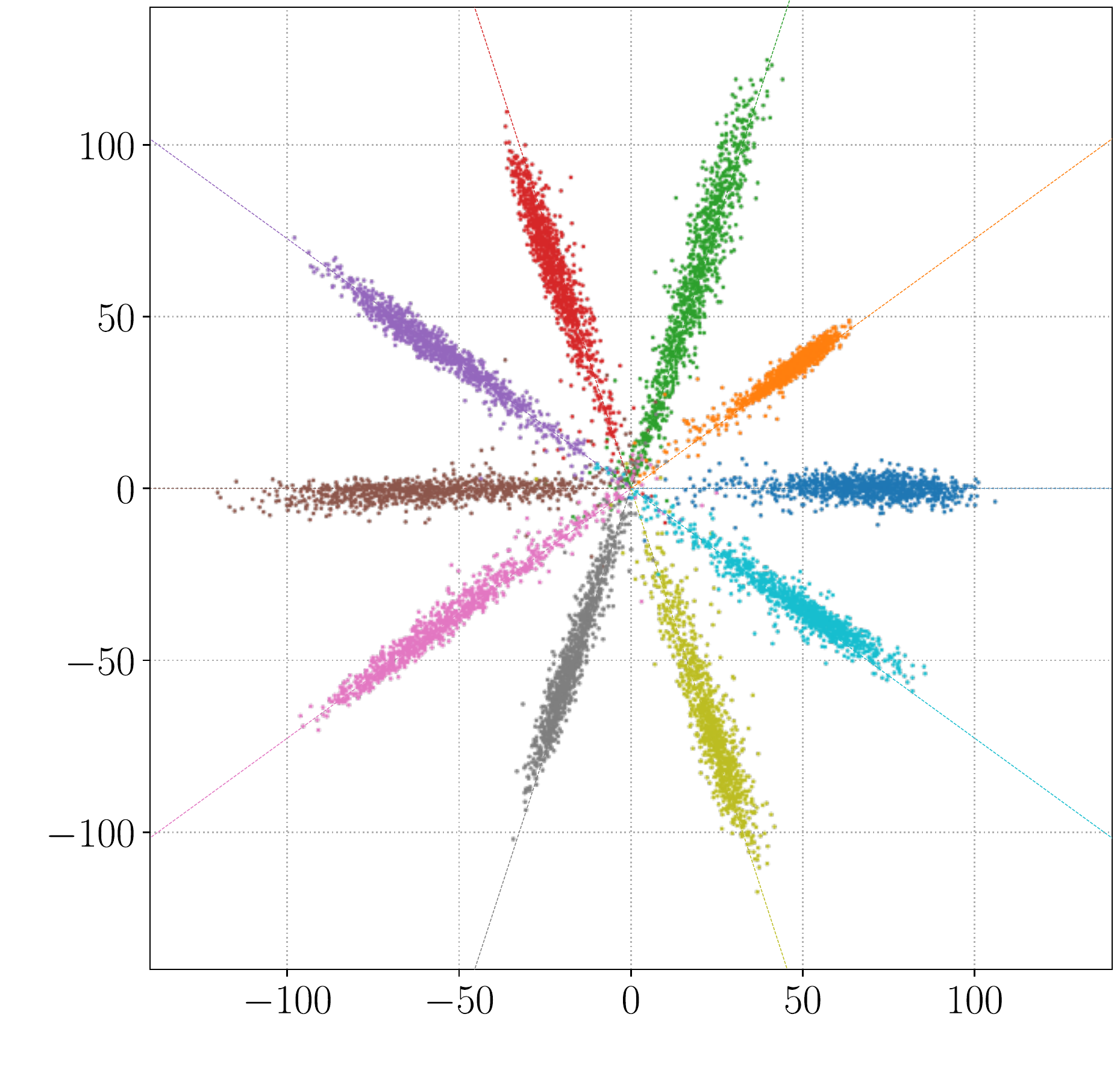}}
\subfloat[]{\label{fig:b}\includegraphics[height=0.181\linewidth,valign=t]{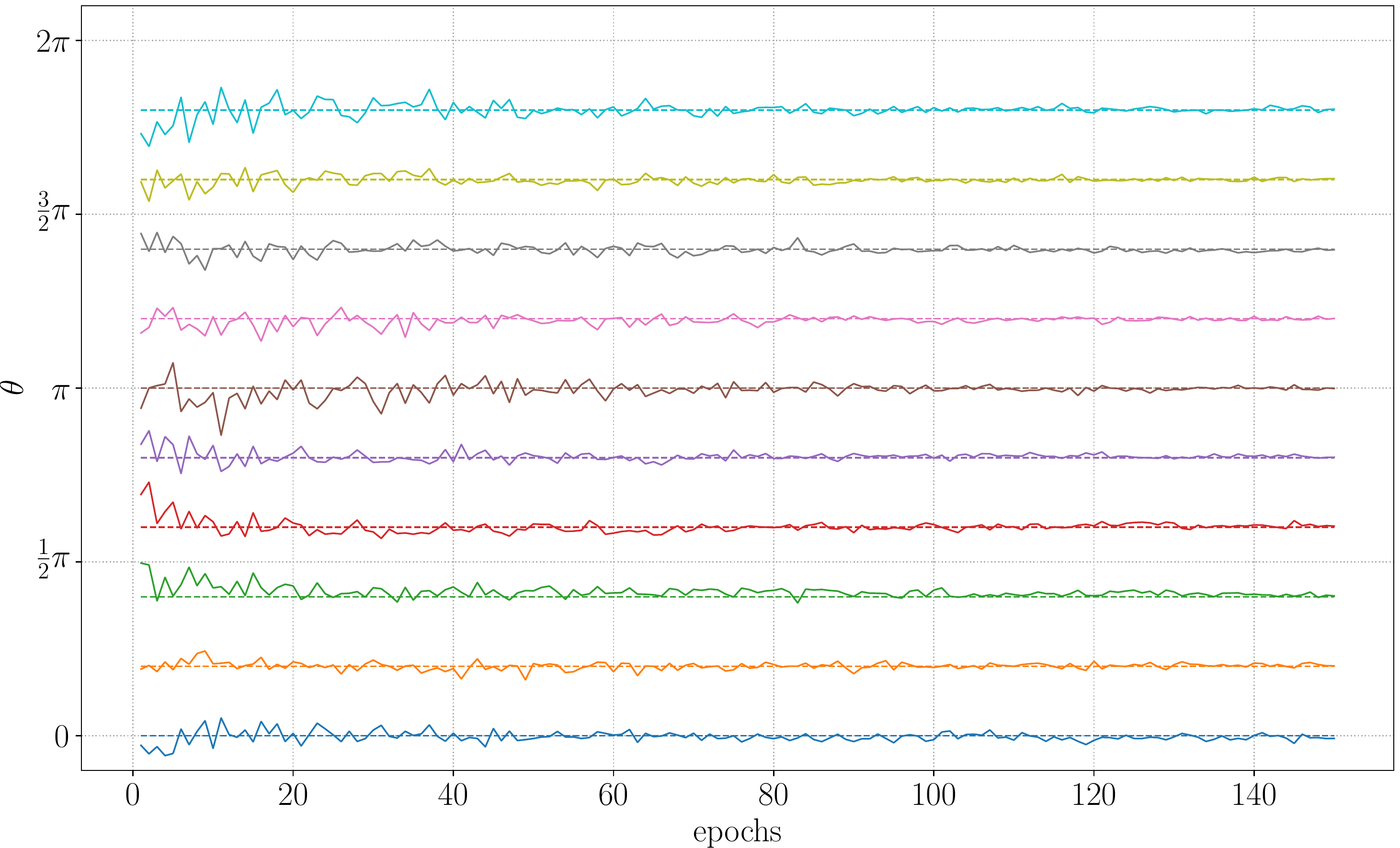}}\hspace{1pt}
\subfloat[]{\label{fig:c}\includegraphics[height=0.181\textwidth,valign=t]{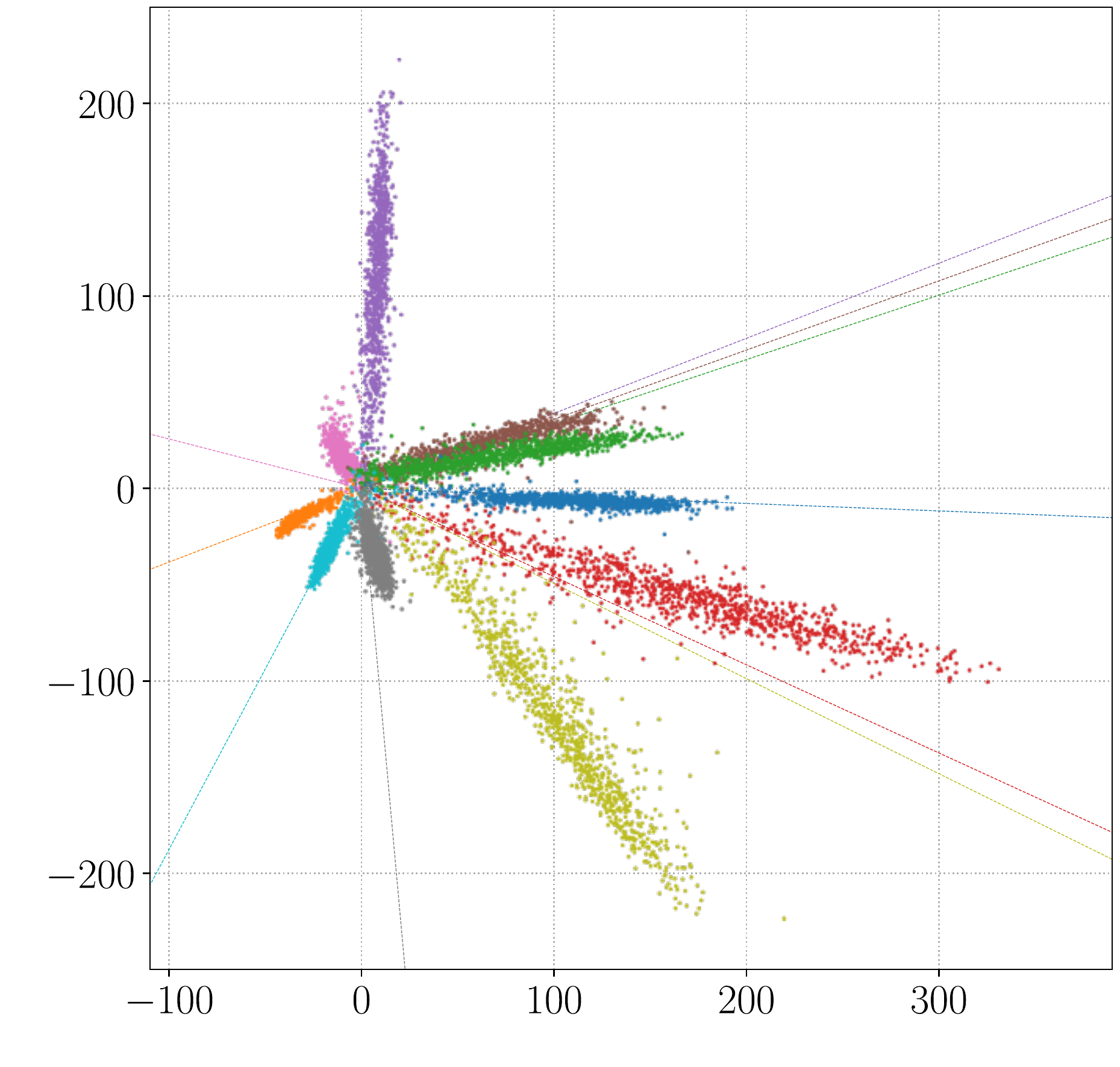}}%
\subfloat[]{\label{fig:d}\includegraphics[height=0.181\textwidth,valign=t]{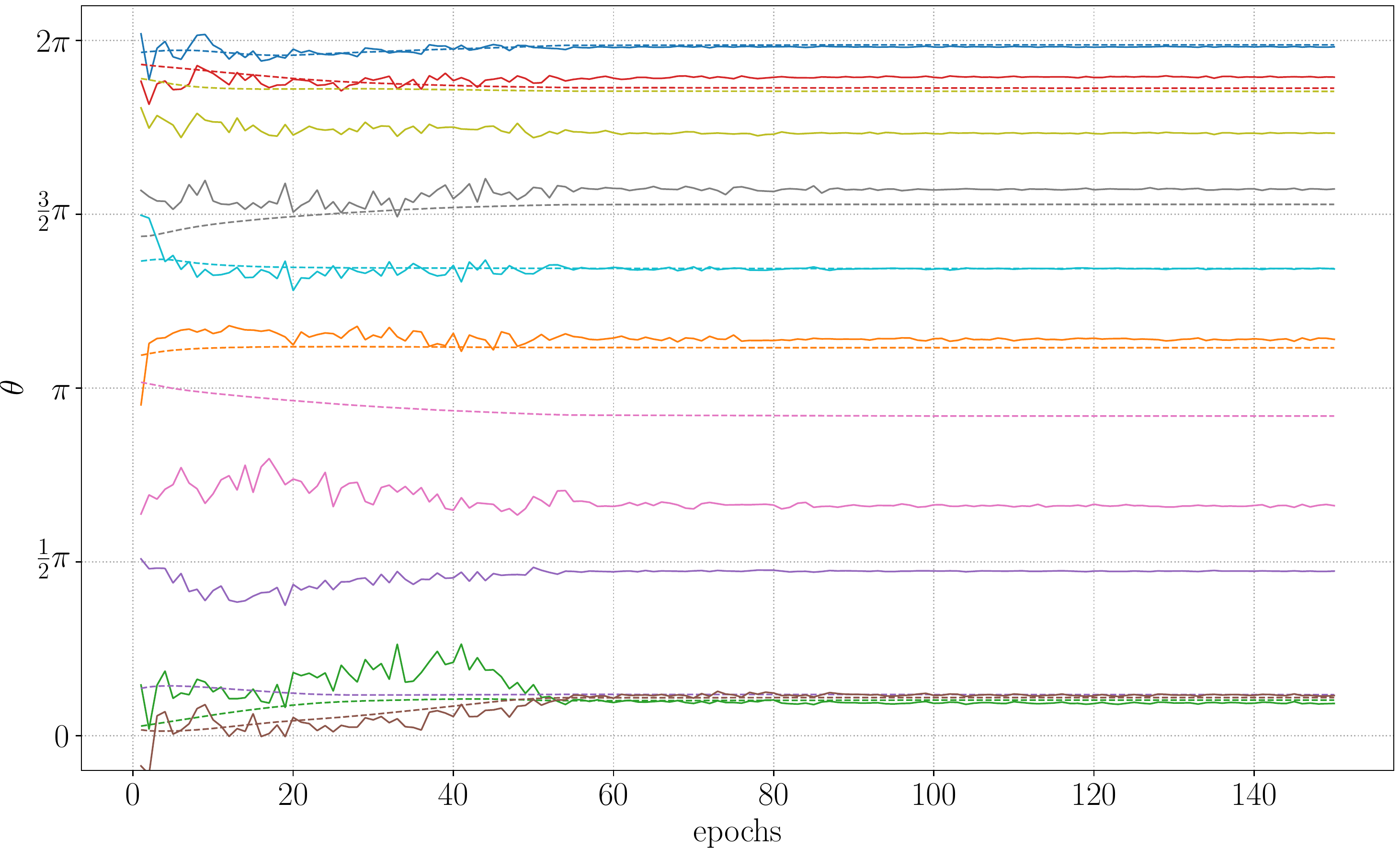}}\hspace{5pt}%
\subfloat{\includegraphics[width=0.033\textwidth,valign=t]{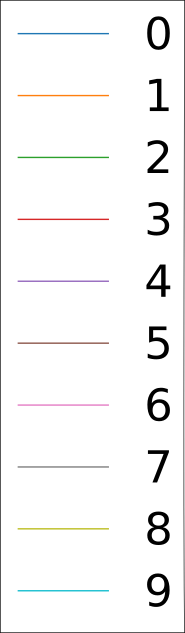}}
\caption{Feature learning on the MNIST dataset in a 2D embedding space. Fig.~(a) and Fig.~(c) show the 2D features learned by RePoNet and by a standard trainable classifier respectively. 
The two methods achieve the same classification accuracy. 
Fig.~(b) and Fig.~(d) show the training evolution of the classifier weights (dashed) and their corresponding class feature means (solid) respectively. Both are expressed according to their angles.
\vspace{-0.4cm}
}
\label{fig:intro_fig_mnist}
\end{figure*}

In this paper we exploit the fact that fixed weights can induce some form of stationarity on the generated features. Indeed, if the parameters $\mathbf{w}_i$ of the classifier in Eq.~\ref{eq_logit} are fixed  (i.e. set as non trainable), \emph{only the feature vector directions} can align toward the classifier weight vector directions and not the opposite. Therefore weights can be regarded as fixed angular references to which features align.

According to this, we provide a precise result on the spatio-temporal statistical properties of the generated features during the learning phase. Supported by the empirical evidence in \cite{hoffer2018fix} we show that not only the final classifier of a DCNN can be set as non trainable but that an appropriate set of values assigned to its weights allows learning a maximally discriminative and strictly stationary embedding while training. That is, the features generated by the Stochastic Gradient Descent (SGD) optimization have constant mean equal to their corresponding class fixed weights. 
Constant known mean implies that features cannot have non-constant trends while learning.
Maximally discriminative features and their stationarity are obtained by setting the classifier weights according to values following a highly symmetrical configuration in the embedding space. 
\\
\indent
DCNN models are typically convergent irrespective of their fixed or not fixed classifier and therefore after that a sufficient learning time has elapsed some form of stationarity in the learned features can still be achieved. However, until that time, it is not possible to  know, where the features will be placed (i.e. projected by the learned model) in the embedding space. 
The main strength of the approach proposed in this paper is that it allows to define (and therefore to know in advance) where the features will be placed  
before starting the learning process.
\\
\indent
Our result can be understood by looking at the basic functionality of the final classifier in a DCNN. The main role of a trainable classifier is to dynamically adjust the decision boundaries to learn class feature representations. When the classifier is set as non trainable this  
dynamic adjustment capability is no longer available and it is automatically demanded to all of the previous layers.  
We demonstrate that this missing functionality can be correctly superseded by the previous layers using a predetermined set of decision boundaries with the highest degree of symmetry in the embedding space so that the generated features are maximally discriminative.
The underlying and largely verified assumption (to be precised later) is that the expressive power of DCNN models is large enough to account for the missing trainable classifier. 
In this novel learning regime the network is basically forced to learn class feature representation into the predetermined symmetric set of subspaces. 

We show that our approach can be theoretically justified and easily implemented by setting the classifier weights to values taken from the coordinate vertices of a regular polytope in the embedding space.
Regular polytopes are the generalization in any number of dimensions of regular polygons and regular polyhedra (i.e. Platonic Solids). Although, there are infinite regular polygons in $\mathbb{R}^2$ and 5 regular polyedra in $\mathbb{R}^3$, there are only three regular polytopes in $\mathbb{R}^d$ with $d \geq 5$, namely the $d$-Simplex, the $d$-Cube and the $d$-Orthoplex. Having different symmetry, geometry and topology, each regular polytope will reflect its properties into the classifier and the embedding space which defines. Fig.~\ref{fig_IntroFig} illustrates the three basic architectures defined by the proposed approach termed Regular Polytope Networks (RePoNet). Fig.~\ref{fig:intro_fig_mnist} provides a first glance at our main result in a 2D embedding space. Specifically, the main evidence from Fig.~\ref{fig:a} and \ref{fig:b} is that the features learned by RePoNet remain aligned with their corresponding fixed weights and maximally exploit the available representation space directly from the beginning of the training phase.

We apply our method to multiple vision datasets showing that it is possible to generate stationary and maximally discriminative features without reducing the generalization performance of DCNN models.

\section{Related Work}
A recent paper \cite{hoffer2018fix} explores the idea of  
excluding the parameters $\mathbf{w}_i$ in Eq.\ref{eq_logit}  from learning. The work shows that a fixed classifier causes little or no reduction in classification performance for common datasets while allowing a noticeable reduction in trainable parameters, especially when the number of classes is large. Setting the last layer as not trainable also reduces the computational complexity for training as well as the communication cost in distributed learning. The described approach sets the classifier with the coordinate vertices of orthogonal vectors taken from the columns of the Hadamard\footnote{The Hadamard matrix is a square matrix whose entries are either +1 or −1 and whose rows are mutually orthogonal.} matrix. Although the work uses a fixed classifier, no mention is made about the fact that the generated features may have stationary properties. A major limitation of this method is that, when the number of classes is greater than the dimension of the feature space, it is not possible to have mutually orthogonal columns and therefore some of the classes are constrained to lie in a common subspace causing a reduction in classification performance. We improve and generalize this work by finding a novel set of unique directions that overcomes the limitations of the Hadamard matrix. 

Some papers exploit Eq.~\ref{eq_logit} to train DCNNs by direct angle optimization \cite{ranjan2017l2,Liu2017CVPR,Liu_2018_CVPR,wang2017normface}. For these papers, among others, the performance is not only dominated by the predicted labels as in \cite{hoffer2018fix}, 
but features are also required to be compact and well separable in terms of their angular directions. 
From a semantic point of view the angle encodes the required discriminative information for class recognition. The wider the angles the better the classes are separated from each other and, accordingly, their representation is more discriminative.  The common idea of these works is that of constraining the features and/or the classifier weights to be unit normalized. The works \cite{liu_2017_coco_v1}, \cite{hasnat2017mises} and \cite{wang2017normface} optimize both features and weights, while the work \cite{ranjan2017l2} normalizes the features only and \cite{Liu2017CVPR} normalizes the weights only. Specifically, \cite{ranjan2017l2} also proposes adding a scale parameter after feature normalization based on the property that increasing the norm of samples can decrease the Softmax loss \cite{yuan2017feature}. 
Despite not being involved in learning discriminative feature, the work \cite{hoffer2018fix} in addition to fixing the classifier, normalizes both the weights and the features and exploits the multiplicative scale parameter. In accordance with \cite{wang2018additive,hoffer2018fix} and \cite{wang2017normface} we found that feature normalization and the multiplicative scale parameter are hard to optimize for general datasets, having a significant dependence on image quality. We follow the work \cite{Liu2017CVPR} that normalizes the classifier weights and sets its biases to zero. This allows to optimize angles, enabling the network to learn angularly distributed features.   

From a statistical point of view normalizing weights is equivalent to consider features distributed on the unit hypersphere according to the von Mises-Fisher distribution \cite{hasnat2017mises}. Under this model each class weight represents the mean of its corresponding features and the scalar factor (i.e. the concentration parameter) is inversely proportional to their standard deviations. Several methods implicitly follow this statistical interpretation \cite{wang2017normface, Liu2017CVPR, wang2018additive, Imprinted_Qi_2018_CVPR, Wu_2018_ECCV} in which the weights act as a summarizer or as parameterized prototype of the features of each  class. Eventually, as conjectured in \cite{wang2017normface} if all classes are well-separated they will roughly correspond to the means of features in each class. 
We improve upon these works by showing that our proposed classifiers produce features exactly centered around to their fixed weights as the training process advances.

While all the above works impose large angular distances between the classes, they provide solutions to enforce such constraint in a local manner without considering global interclass separability and intraclass compactness. For this purpose, very recently the work \cite{HypersphericalEnergy2018} adds a regularization loss to specifically force the classifier weights to be far from each other in a global manner. The work draws inspiration from a well-known problem in physics -- the Thomson problem \cite{thomson1904xxiv}, where given $K$ charges confined to the surface of a sphere, one seeks to find an arrangement of the charges which minimizes the total electrostatic energy. Electrostatic force repels charges each other inversely proportional to their mutual distance. In \cite{HypersphericalEnergy2018} global equiangular features are obtained by adding to the standard categorical Cross-Entropy loss a further loss inspired by the Thomson problem.  
We follow a similar principle for global separability and compactness by considering that minimal energies are often concomitant with special geometric configurations of charges that recall the geometry of Platonic Solids in high dimensional spaces \cite{batle2016generalized}.

\section{Main Contributions}
Our technical contributions can be summarized as follows:
(1) We generalize the concept of fixed classifiers. (2) We show that they can generate stationary and maximally discriminative features at training time.
(3) We provide a theoretical model that justifies our result.

\section{Regular Polytopes and Maximally Discriminative Stationary Embeddings}
We are basically concerned with the following question: \emph{How should the non trainable weights be distributed in the embedding space such 
that they generate stationary and maximally discriminative features?}

Let $\mathbb{X} = \{(x_i, y_i)\}_{i=1}^{N}$ be the training set containing $N$ samples, where $x_i$ is the raw input to the DCNN and $y_{i} \in \{1,2,\cdots,K\}$ is the label of the class that supervises the output of the DCNN. Then, the Cross Entropy loss can be written as:
\begin{align}
\mathcal{L}=-\frac{1}{N}\sum_{i=1}^{N} \log\Bigg( \frac {\exp({\mathbf{w}_{y_i}^{\top}\mathbf{f}_{i}+\mathbf{b}_{y_i}})} {\sum_{j=1}^{K}\exp({\mathbf{w}_{j}^{\top}\mathbf{f}_{i} + \mathbf{b}_{j}})} \Bigg), 
\label{softmax_loss}
\end{align}
where $\mathbf{W}=\{ \mathbf{w}_j \}_{j=1}^{K}$ are the classifier weight vectors for the $K$ classes. 
Following the discussion in \cite{wang2017normface} we normalize the weights and zero the biases ($\hat{\mathbf{w}}_j = \frac{\mathbf{w}_j}{||\mathbf{w}_j||}$, $\mathbf{b}_j=0$) to directly optimize angles, enabling the network to learn angularly distributed features.

Angles therefore encode the required discriminative information for class recognition and the wider they are, the better the classes are represented. As a consequence, the representation in this case is maximally discriminative when features are distributed at \emph{equal angles} maximizing the available space. 

If we further consider, the feature vector parametrized by its unit vector as $\mathbf{f}_i=\kappa_i \, \hat{\mathbf{f}}_i$ where $\kappa_i=||\mathbf{f}_i||$ and $\hat{\mathbf{f}}_i = \frac{\mathbf{f}_i}{||\mathbf{f}_i||}$ then Eq.\ref{softmax_loss} can be rewritten as: 
\begin{align}
\mathcal{L}=-\frac{1}{N}\sum_{i=1}^{N}\log\Bigg( \frac {\exp ( { \kappa_i \hat{\mathbf{w}}_{y_i}^{\top}\hat{\mathbf{f}}_i })} {\sum_{j=1}^{K}\exp({ \kappa_i \hat{\mathbf{w}}_{j}^{\top}\hat{\mathbf{f}}_i })} \Bigg)  
\label{softmax_loss_von}
\end{align}
The equation above can be interpreted as if $N$ realizations from a set of $K$ von Mises-Fisher distributions with different concentration parameters $\kappa_i$ are passed through the Softmax function. 
The probability density function of the von Mises-Fisher distribution for the random $d$-dimensional unit vector $\hat{\mathbf{f}}$ is given by:
$
P(\hat{\mathbf{f}} ;\hat{\mathbf{w}},\kappa ) \propto \exp  \big ( {\kappa \hat{\mathbf{w}}^{\top} \hat{\mathbf{f}} } \big ) 
\label{eq_vonMisesFisher}
$
where $ \kappa \geq 0$.
Under this parameterization $\hat{\mathbf{w}}$ is the mean direction on the hypersphere and $\kappa$ is the concentration parameter. The greater the value of $\kappa$ the higher the concentration of the distribution around the mean direction $\hat{\mathbf{w}}$. The distribution is unimodal for $\kappa>0$ and is uniform on the sphere for $\kappa=0$. The loss in Eq.~\ref{softmax_loss_von} optimizes for large values of $\kappa$ and small angles providing intraclass compactness.

As with this formulation each weight vector is the mean direction of its associated features on the hypersphere, equiangular features maximizing the available space can be obtained by arranging accordingly their corresponding weight vectors around the origin. This problem is equivalent to distributing points uniformly on the sphere and is a well-known geometric problem, called Tammes problem  \cite{tammes1930origin} which is a generalization of the physic problem firstly addressed by Thomson \cite{thomson1904xxiv}. 
In 2D the problem is that of placing $K$ points on a circle so that they are as far as possible from each other. In this case the optimal solution is that of placing the points at the vertices of a regular $K$-sided polygon. 

The 3D analogous of regular polygons are Platonic Solids. However, the five Platonic solids are not the unique solutions of the Thomson problem. In fact, only the tetrahedron, octahedron and the icosahedron are the unique solutions for $K = 4$, $6$ and $12$ respectively. For $K = 8$: the cube is not optimal in the sense of the Thomson problem. This means that the energy stabilizes at a minimum in configurations that are not symmetric from a geometric point of view. The unique solution in this case is provided by the vertices of an irregular polytope \cite{bagchi1997stay}.  

The non geometric symmetry between the locations causes the global charge to be different from zero. Therefore in general, when the number of charges is arbitrary, their position on the sphere cannot reach a configuration for which the global charge vanishes to zero. A similar argument holds in higher dimensions for the so called generalized Thomson problem \cite{batle2016generalized}. According to this, we argue that, \emph{the geometric limit to obtain a zero global charge in the generalized Thomson problem is equivalent to the impossibility to obtain a maximally discriminative classifier for an arbitrary number of classes}. 

However, since the classification problem it is not confined in a three dimensional space, our approach addresses this irregularity 
by \emph{selecting the appropriate dimension of the embedding space so as to have access to symmetrical fixed classifiers directly from regular polytopes}.  
In dimensions 5 and higher, there are only three ways to do that (See Tab.~\ref{tab_polytopes}) and they involve the symmetry properties of the three well known regular polytopes available in high dimensional space \cite{coxeter1963regular}. These three special classes exist in every dimensionality and are: the $d$-Simplex, the $d$-Cube and the $d$-Orthoplex. 
In the next paragraphs the three fixed classifiers derived from them are presented.
\begin{table}[t]
\centering
\begin{tabular}{@{}l|lllll@{}}
\toprule
\toprule
Dimension $d$ & $1$ & $2$ & $3$ & $4$ & $\geq 5$ \\ \midrule
Number of Regular Polytopes & $1$ & $\infty$ & $5$ & $6$ & $\mathbf{3}$ \\ 
\bottomrule
\bottomrule
\end{tabular}
\caption{Number of regular Polytopes as dimension $d$ increases. \vspace{-0.2cm}}
\label{tab_polytopes}
\end{table}

\paragraph{The $d$-Simplex Fixed Classifier.}
In geometry, a simplex is a generalization of the notion of a triangle or tetrahedron to arbitrary dimensions. Specifically, a $d$-simplex is a $d$-dimensional polytope which is the convex hull of its $d + 1$ vertices. A regular $d$-simplex may be constructed from a regular $(d-1)$-simplex by connecting a new vertex to all original vertices by the common edge length. According to this, the weights for this classifier can be computed as: 
\begin{align}
\mathbf{W}_S=\Big \{e_1,e_2,\dots,e_{d-1}, \alpha \sum_{i=1}^{d-1} e_i \Big \}
\nonumber
\end{align}
where $\alpha=\frac{1-\sqrt{d+1}}{d}$ and $e_i$ with $i \in \{1,2, \dots, d-1\}$ denotes the standard basis in $\mathbb{R}^{d-1}$. The final weights will be shifted about the centroid and normalized.
The $d$-Simplex fixed classifier defined in an embedding space of dimension $d$ can accomodate a number of classes equal to its number of vertices:
\begin{align}
K=d+1.
\label{eq_ksimplex}
\end{align}
This classifier has the largest number of classes that can be embedded in $\mathbb{R}^d$ such that their corresponding class features are equidistant from each other. It can be shown that the angle subtended between \emph{any} pair of weights is equal to: 
\begin{align}
\theta_{\mathbf{w}_i,\mathbf{w}_j}=\arccos\bigg(-\frac{1}{d}\bigg) \quad \forall i,j \in \{ 1, 2, \dots, K\} : i \neq j.
\label{eq_dsimplex_angle}
\end{align}

\paragraph{The $d$-Orthoplex Fixed Classifier.}
This classifier is derived from the $d$-Ortohoplex (or Cross-Polytope) regular polytope that is defined by the convex hull of points, two on each Cartesian axis of an Euclidean space, that are equidistant from the origin. The weights for this classifier can therefore defined as:
$$\mathbf{W}_O = \{ \pm e_1,\pm e_2,\dots,\pm e_d \}.$$
Since it has $2d$ vertices the derived fixed classifier can accommodate in its embedding space of dimension $d$:
\begin{align}
K=2d
\label{eq_kortho}
\end{align}
different classes.  
Each vertex is connected to other $d-1$ vertices and the angle between connected vertices is 
\begin{align}
\theta_{\mathbf{w}_i,\mathbf{w}_j}= \frac{\pi}{2} \quad \forall  \, i,j  \in \{ 1, 2, \dots, K\}  : j \in C(i)
\label{eq_dortho_angle}
\end{align}
Where each $j \in C(i)$ is a connected vertex and $C$ is the set of connected vertices defined as ${C}(i)=\{ j:(i,j) \in {E}\}$. $E$ is the set of edges of the graph ${G}=(\mathbf{W}_O,{E})$.
The $d$-Orthoplex is the dual polytope of the $d$-Cube and vice versa (i.e. the normals of the $d$-Orthoplex faces correspond to the the directions of the vertices of the $d$-Cube).

\paragraph{The $d$-Cube Fixed Classifier.}
The $d$-Cube (or Hypercube) is the regular polytope formed by taking two congruent parallel hypercubes of dimension $(d-1)$ and joining pairs of vertices, so that the distance between them is $1$. A $d$-cube of dimension 0 is one point.
The fixed classifier derived from the $d$-Cube
is constructed by creating a vertex for each binary number in a string of $d$ bits. Each vertex is a $d$-dimensional boolean vector with binary coordinates $-1$ or $1$. Weights are finally obtained from the normalized vertices:
$$ 
\mathbf{W}_c = \left  \{ \mathbf{w} \in \mathbb {R}^{d}:  \left [-\frac{1}{\sqrt{d}},\frac{1}{\sqrt{d}} \right ] ^d \right \}.
$$ 
The $d$-Cube can accommodate: 
\begin{align}
K=2^d
\label{eq_kcube}
\end{align}
classes. The vertices are connected by an edge whenever the Hamming distance of their binary numbers is one therefore forming a $d$-connected graph.
The angle of a vertex with its connected vertices is:
\begin{align}
\theta_{\mathbf{w}_i,\mathbf{w}_j} = \arccos\bigg(\frac{d-2}{d}\bigg),  \forall  \, i,j  \in \{ 1, \dots, K\}  : j \in C(i)
\label{eq_dcube_angle}
\end{align}
where $C(i)$ is the set of vertices connected to the vertex $i$.
\paragraph*{}
Given a classification problem with $K$ classes, the three RePoNet fixed classifiers can be simply instantiated by defining a non trainable fully connected layer of dimension $d$, where $d$ is computed from Eqs.~\ref{eq_ksimplex}, \ref{eq_kortho} and \ref{eq_kcube} as summarized in the following table:  
\vspace{-0.2cm}
\begin{table}[h]
\begin{tabular}{@{}c|ccl@{}}
\toprule
\toprule
RePoNet  & $d$-Simplex & $d$-Cube & $d$-Orthoplex \\ \midrule
Layer dim. & $d=K-1$ & $d = \lceil \log_2(K) \rceil$ & $d = \big \lceil \frac{K}{2} \big \rceil$ \\ \bottomrule \bottomrule
\end{tabular}
\label{tab_dims}
\vspace{-0.3cm}
\end{table}
\FloatBarrier

\begin{figure}[h]
\centering
\includegraphics[width=0.99\columnwidth]{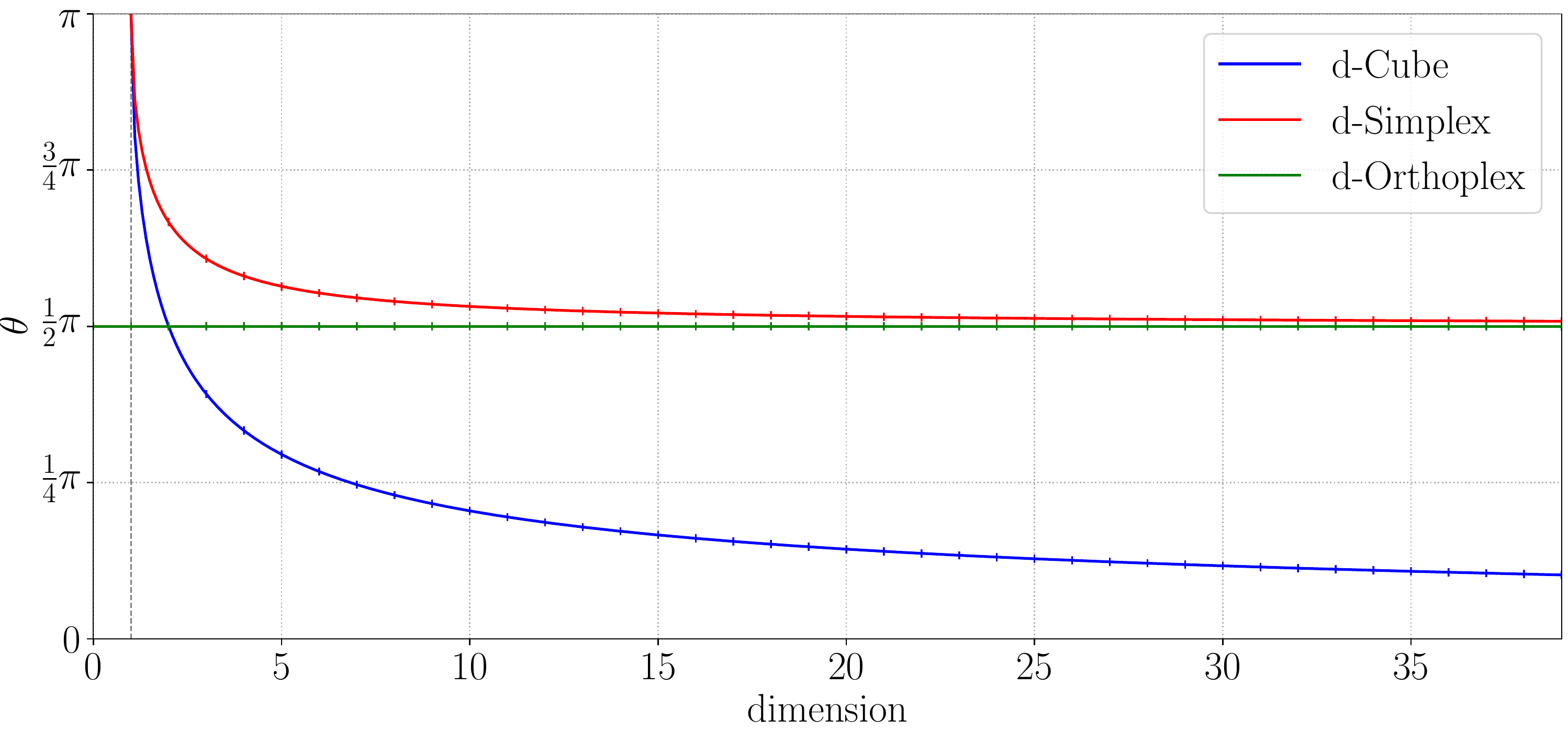}
\caption{The angular space defined by RePoNet classifiers. Curves represent the angle between a weight and its connected weights as the dimension of the embedding space increases. The angle between class features follows the same trend. }
\label{fig:anglefcn}
\end{figure}

Fig.~\ref{fig:anglefcn} shows the angle between a weight and its connected weights computed from Eqs.~\ref{eq_dsimplex_angle}, \ref{eq_dortho_angle} and \ref{eq_dcube_angle} as the dimension of the embedding space increases. Having the largest angle between its weights, the $d$-Simplex fixed classifier achieves the best intra-class separability. However, as the embedding space dimension increases, its angle tends towards $\pi/2$. Therefore the largest the dimension of the space the more it becomes similar to the $d$-Orthoplex classifier. The main difference between the two classifiers is in their neighbors connectivity. The different connectivity of the three regular polytope classifiers has a direct influence on the evaluation of the loss. In the case of the $d$-Simplex classifier, all the summed terms in the loss of Eq.~\ref{softmax_loss_von} have always comparable magnitudes in a mini batch. 

The $d$-Cube classifier has the most compact feature embedding and the angle between each weight and its $d$ neighbors decreases as the dimension increases. Accordingly, it's the hardest to optimize. 

\section{Theoretical Analysis}
\newtheorem{theorem}{Theorem}
\theoremstyle{definition}
\newtheorem{definition}{Definition}
The joint density $P(\mathbf{f}_1, \mathbf{f}_2, \dots, \mathbf{f}_K)$ specified by a learned DCNN model encapsulates the dependence among the features of the $K$ classes. In general, there is a large number of ways about the form in which such dependencies can be learned, but there are some simple forms which accurately describe the way in which features are learned by the proposed approach.
\begin{figure}[t]
\centering
\includegraphics[width=0.65\columnwidth]{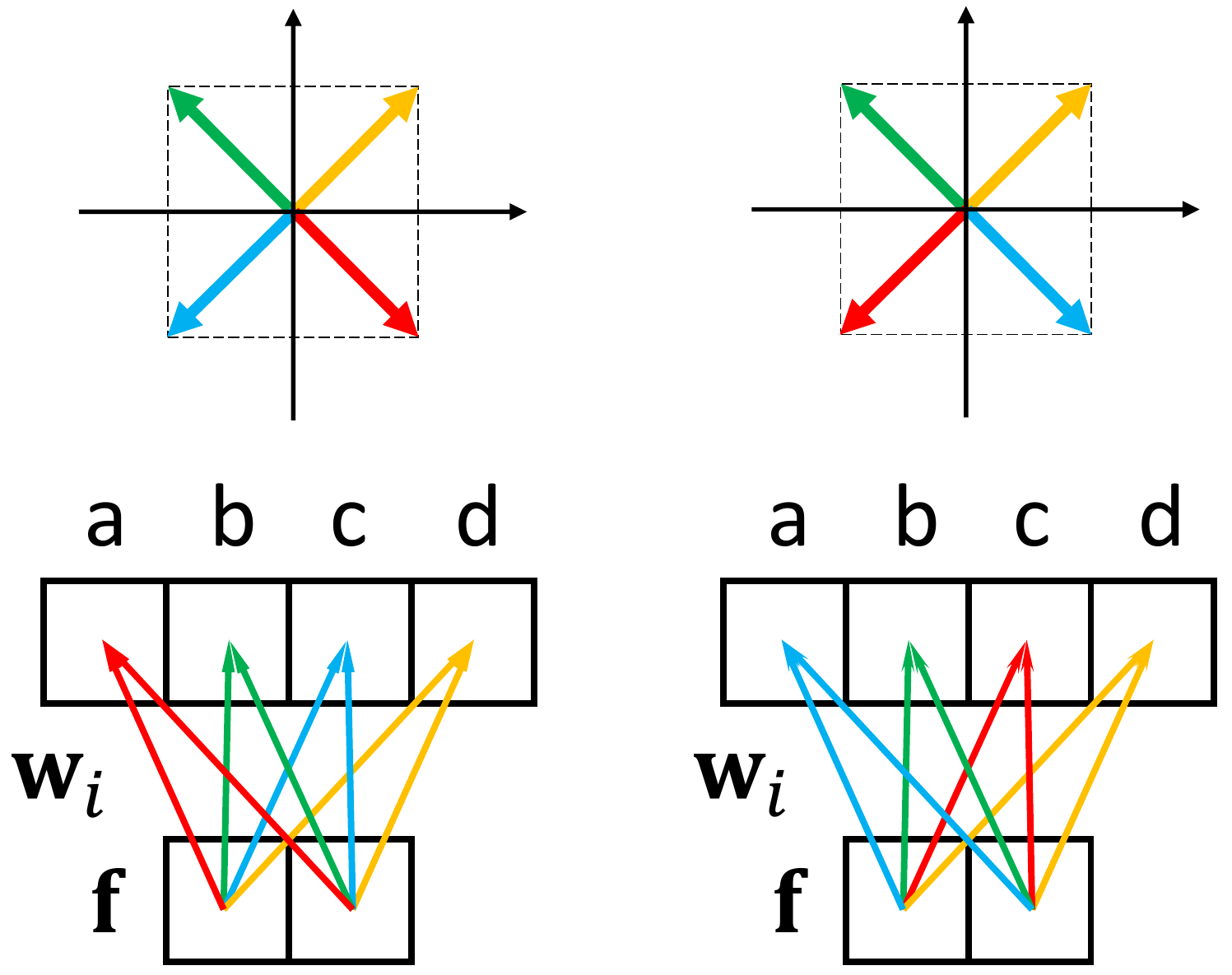}
\caption{A fixed classifier in a four class 2D feature space scenario. Exchanging the red and blue classifier fixed weights is equivalent to exchange the labels $a$ and $c$ and their associated features. However, due to the symmetry of 4-sided regular polygon (\emph{top}) the dependency between the class features remains the same.
}
\label{fig_exchangeability_weights}
\end{figure}
Suppose that, in considering the joint distribution above, the labels identifying the individual features
are uninformative, in the sense that the information the $\mathbf{f}_i$ 
provide is independent of the order in which the labels $y_i$ are presented to the output of the classifier. This order independence is captured by the property of \emph{Exchangeability} (or symmetrical dependency) \cite{aldous1985exchangeability,Bernardo1996}. 
With reference to Fig.~\ref{fig_exchangeability_weights}, exchanging the weights $\mathbf{w}_i$ is equivalent to exchange the associated labels and consequently their associated learned features. However, due to the symmetrical  configuration of the weights, the form that shapes the dependency between the features does not change. More formally:
\theoremstyle{definition}
\begin{definition}{Exchangeability.}
Let $\pi$ denote an arbitrary permutation on $n$ elements (that is, a function that maps $ \{1, \dots , n \} $ onto itself). The finite sequence $X_1 , \dots , X_n $ is said to be exchangeable if the joint distribution of the permuted random vector $( X_{ \pi(1) } , \dots , X_{\pi(n)} )$ is the same no matter which  is chosen. The infinite sequence $X_1 , X_2 , \dots $ is said to be exchangeable if any finite subsequence is exchangeable.
\label{Exchangeability}
\end{definition}
Exchangeable sequences are basically random sequences that are invariant to the class of transformations representing permutations. This property is directly related to strict stationarity of random sequences.
\theoremstyle{definition}
\begin{definition}{Stationarity.}
The sequence $X_1, X_2, \dots $ is said to be stationary if, for a fixed $k \geq 0$,
the joint distribution of $(X_i, \dots , X_{i+k})$ is the same no matter what positive value of $i$ is chosen.
\end{definition}

Under the assumption (empirically confirmed in Sec.~\ref{sec_Exchangeability_Assumption}) that a DCNN with a fixed regular polytope classifier has sufficient expressive power to learn any permutation of its labels, then the generated features are stationary. 

\begin{theorem}
A fixed classifier DCNN with weights set to values taken from the coordinate vertices of a regular polytope generates at training time stationary features.
\end{theorem}
\begin{proof}
It directly follows from the Def.~\ref{Exchangeability} that an exchangeable process is stationary.
\end{proof}

\section{Experimental Results}

We used standard datasets to evaluate both the correctness and the performance of our approach.  
All the experiments are conducted with the well known MNIST, FashionMNIST \cite{FashionMNIST2017}, EMNIST \cite{EMNIST17}, CIFAR-10 and CIFAR-100 \cite{cifar-10}, datasets. We train all the networks using Adam \cite{kingma2014adam}, and the network initialization follows \cite{KaimingHe15}. All the networks use Batch Normalization \cite{Ioffe17} and ReLU if not otherwise specified. 
MNIST and FashionMNIST contain $50,000$ training images and $10,000$ test images. The images are in grayscale and the size of each image is $28 \times 28$ pixels. There are 10 possible classes of digits and clothes respectively. The EMNIST dataset (balanced split) holds $112,800$ training images, $18,800$ test images and has $47$ classes including lower/upper case letters and digits. CIFAR-10 contains $50,000$ training images and $10,000$ test images. The images are in color and have a resolution of $32 \times 32$ pixels. There are 10 classes of various animals and vehicles. CIFAR-100 holds the same number of images of same size, but contains 100 different classes.

\subsection{Exchangeability Assumption}
\label{sec_Exchangeability_Assumption}
In order to provide empirical support for the  theoretical assumption of exchangeability made in this paper we ran a set of experiments to specifically evaluate the expressive power of our approach to learn any permutation of its labels. According to this, we generate random permutations of the labels and a new model is learned for each generated permutation. Since fixed classifiers cannot rely on an adjustable set of subspaces for class feature representation we want to test if some permutations are harder then others for our proposed method. The presence of such hard permutations not only would preclude the underlying exchangeability property but it would also preclude the general applicability of our method. 
The standard trainable classifier does not suffer this problem since it is not forced to learn feature representations into predetermined set of subspaces. 
When features cannot be well separated a trainable classifier can rearrange its (randomly initialized) feature subspaces directions so that the previous convolutional layers can better disentangle the non-linear interaction between complex data patterns. 

\begin{figure}[t]
\centering
\includegraphics[width=0.99\columnwidth]{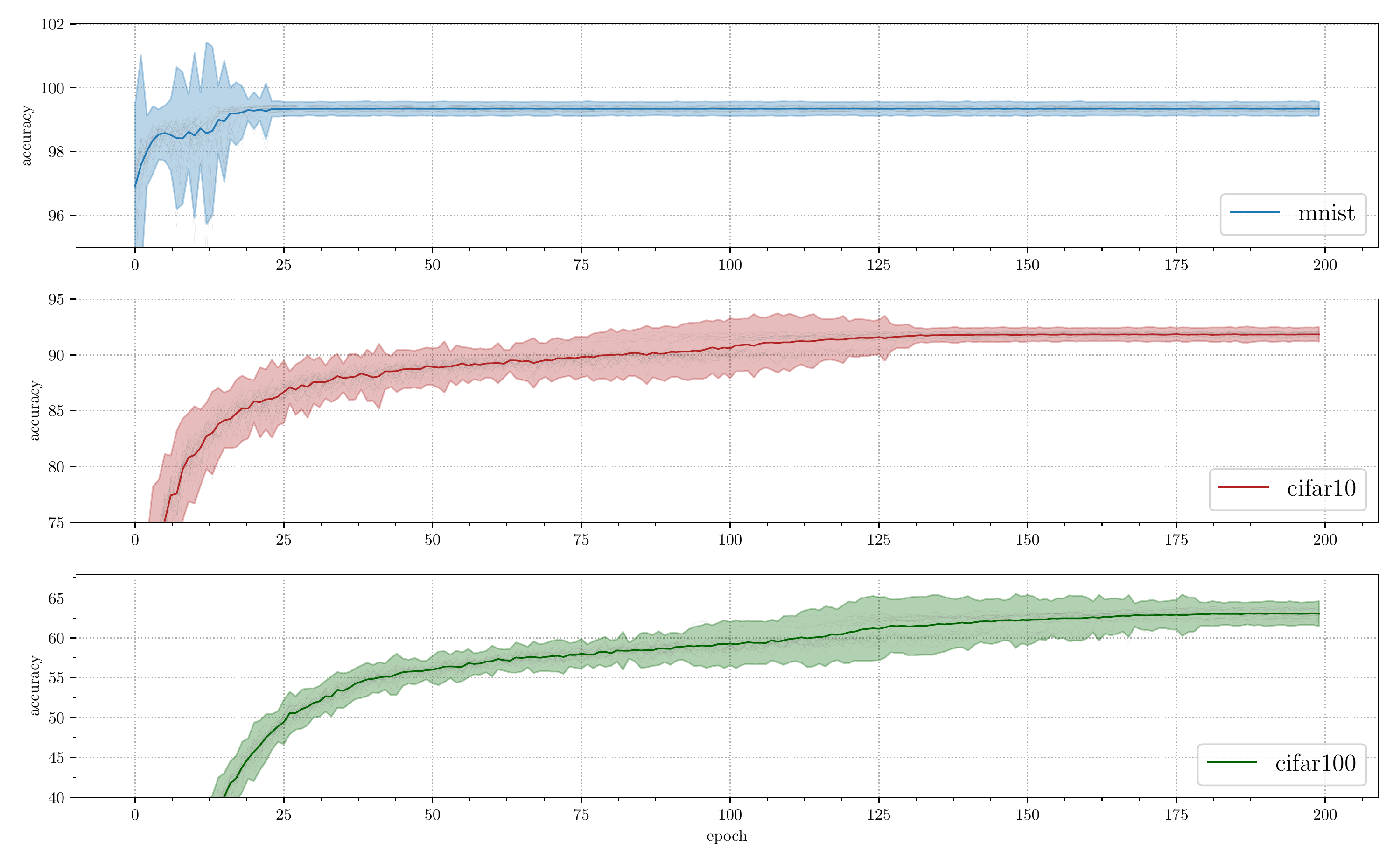}
\vspace{-0.2cm}
\caption{Average accuracy curves and confidence interval computed from the MNIST, CIFAR10 and CIFAR100 datasets under different random permutations of the labels. 
\vspace{-0.3cm}
}
\label{fig_exchangeability}
\end{figure}

Fig.~\ref{fig_exchangeability} shows the mean and the $95\%$ confidence interval computed from the accuracy curves of the learned models. To provide further insights about this analysis, 20 out of 500 accuracy curves computed for each datasets are also shown.  Specifically, the evaluation is performed on three different datasets with an increasing level of complexity (i.e MNIST, CIFAR-10 and CIFAR-100).  
All the models are trained for 200 epochs to make sure that the models trained with CIFAR-100 achieve convergence. 

In order to address the most severe possible outcomes that may happen, for this experiment we used the $d$-Cube fixed classifier. Being the hardest to optimize, this experiment can be regarded as a worst case analysis scenario for our method. As evidenced from the figure, the performance is substantially insensitive to both permutations and datasets. The average reduction in performance at the end of the training process is substantially negligible and the confidence intervals reflect the complexity of the datasets. Although the space of permutations cannot be exhaustively evaluated even for a small number of classes, we have achieved proper convergence for the whole set of 1500 learned models. The experiment took 5 days on a Nvidia DGX-1 (8 Tesla V100).

\begin{figure}[t]
\vspace{-0.2cm}
\centering
\hspace{-0.6cm}
\subfloat[]{\label{fig_even_odd:a}
\includegraphics[height=0.48\linewidth,valign=t]{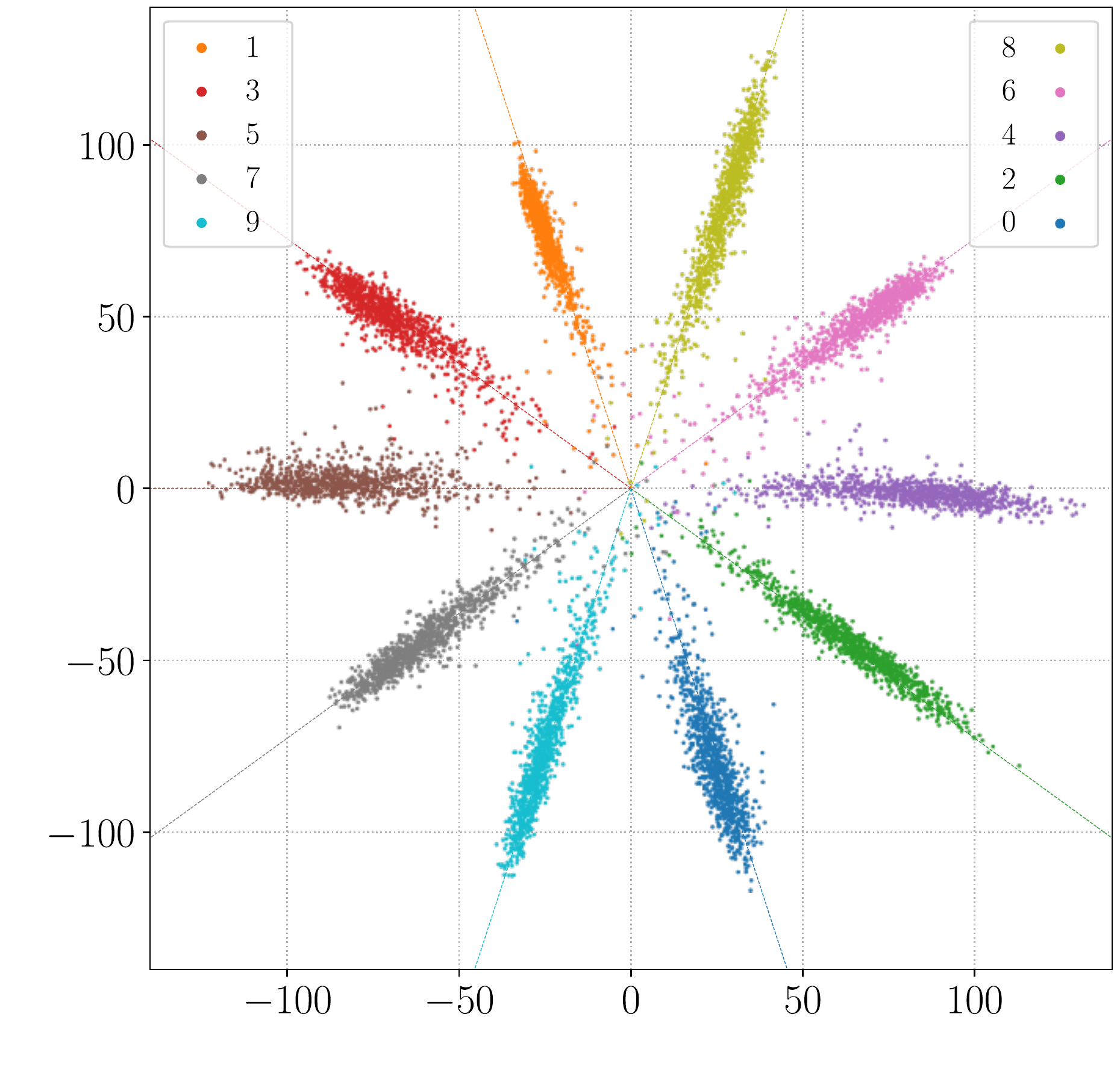}
}
\subfloat[]{\label{fig_CIFAR10:b}
\includegraphics[height=0.48\linewidth,valign=t]{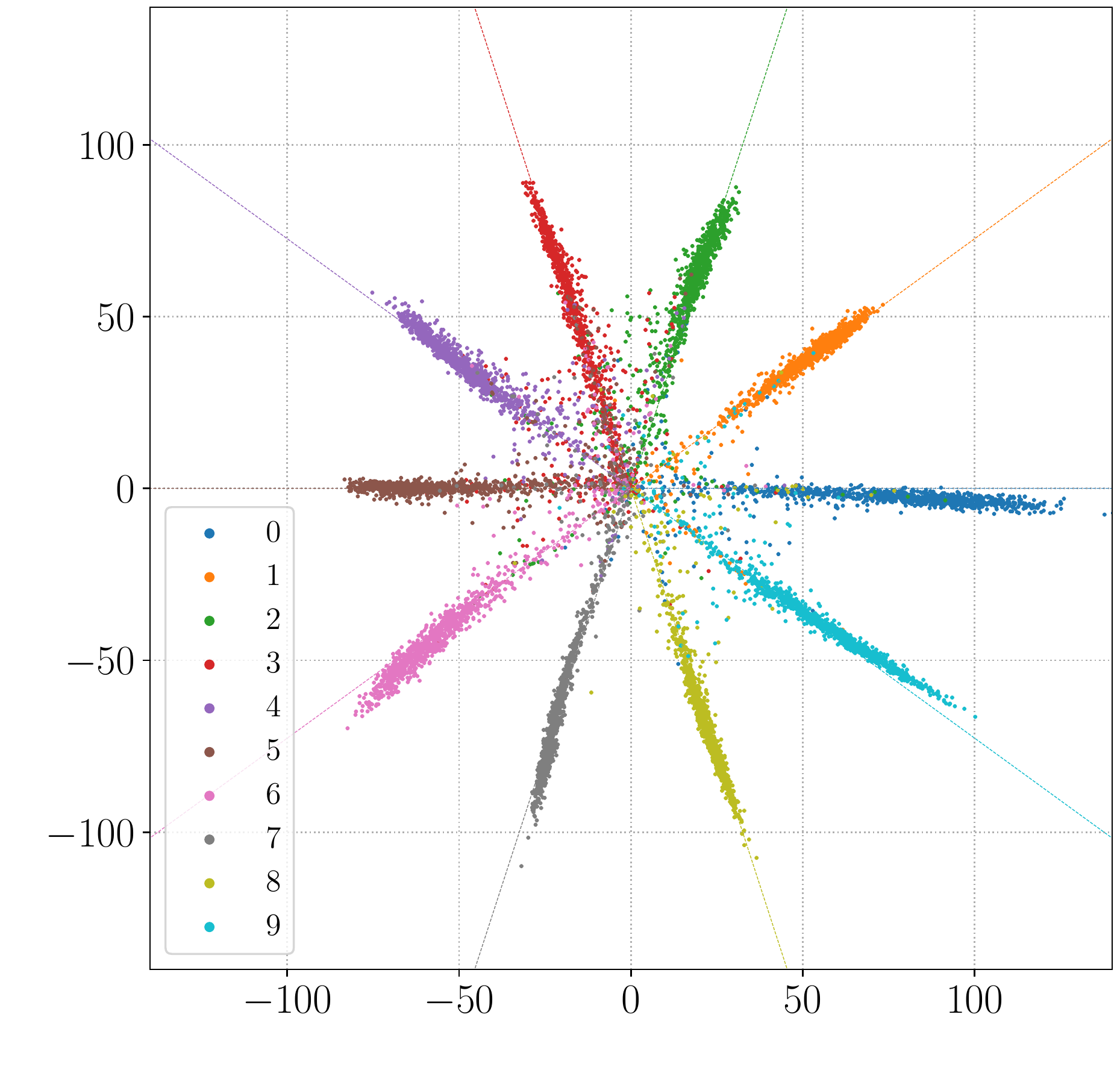}
}
\caption{ The distribution of features learned using a 10-sided regular polygon. \emph{(a)}: A special permutation of classes is shown in which the MNIST even and odd digits are placed in the positive and negative half-space of the abscissa respectively. \emph{(b)}: The features learned using the CIFAR10 dataset.
\vspace{-0.3cm}
}
\label{fig_even_odd}
\end{figure}

We further report qualitative results on the exchangeability property and on the capability of deciding where the features of each class will be projected before starting the training phase. 
Fig.~\ref{fig_even_odd:a} shows features learned in a $k$-sided polygon (2d embedding space) on the MNIST dataset. In particular the model is learned with the permutation (manually selected) of the labels that places even and odd digits features respectively on the positive and negative half space of the abscissa. 
Fig.~\ref{fig_CIFAR10:b} shows the features of CIFAR10 learned with a similar $10$-sided-polygon. It can be noticed that features are distributed following the polygonal pattern shown in Fig.~\ref{fig_even_odd:a}.    

\setlength{\extrarowheight}{2pt} 
\begin{table*}
	\centering
	\footnotesize 
\begin{tabular}{|c|c|c|c|c|c|c|c|}
  \hline

  \multirow{2}{*}{ \diagbox[dir=NW]{\raisebox{0ex}{Method}}{Dataset}} &
  \multicolumn{2}{c|}{ \thead{CIFAR-10 \\[-2px] \scriptsize ($K=10$)}  } & \multicolumn{2}{c|}{ \thead{CIFAR-100 \\[-2px] \scriptsize ($K=100$)}}  &
  \thead{MNIST \\[-2px] \scriptsize ($K=10$)}  & \thead{EMNIST \\[-2px] \scriptsize ($K=47$) } & \thead{ FashionMNIST \\[-2px] \scriptsize ($K=10$) }\\

  \cline{2-8}
  & VGG13 & VGG19 & VGG13 & VGG19 & \multicolumn{3}{c|}{LeNet++}\\

  \hline\hline
  RePoNets $K$-sided-polygon             & $90.57_{d=2}$     & $91.04_{d=2}$     & $35.80_{d=2}$     & $37.65_{d=2}$     & $99.24_{d=2}$     & $72.81_{d=2}$     & $92.48_{d=2}$ \\
  Hadamard fixed classifier \cite{hoffer2018fix}      & $19.69_{d=2}$     & $19.59_{d=2}$     & $\ \,1.69_{d=2}$  & $\ \,1.75_{d=2}$  & $21.14_{d=2}$     & $\ \,4.12_{d=2}$  & $19.89_{d=2}$ \\ \hline
  Baseline (learned classifier)                & $90.82_{d=2}$     & $91.07_{d=2}$     & $37.47_{d=2}$     & $35.83_{d=2}$     & $99.21_{d=2}$     & $73.08_{d=2}$     & $92.79_{d=2}$ \\

  \hline\hline
  RePoNets $d$-Cube                      & $92.17_{d=4}$     & $92.23_{d=4}$     & $62.00_{d=7}$     & $64.02_{d=7}$     & $99.49_{d=4}$     & $88.00_{d=6}$     & $93.72_{d=4}$ \\
  Hadamard fixed classifier \cite{hoffer2018fix}      & $37.09_{d=4}$     & $36.84_{d=4}$     & $\ \,5.31_{d=7}$  & $\ \,5.48_{d=7}$  & $41.45_{d=4}$     & $15.74_{d=6}$     & $38.06_{d=4}$ \\ \hline
  Baseline (learned classifier)                & $92.11_{d=4}$     & $92.36_{d=4}$     & $64.41_{d=7}$     & $66.30_{d=7}$     & $99.56_{d=4}$     & $86.66_{d=6}$     & $94.04_{d=4}$ \\

  \hline\hline
  RePoNets $d$-Orthoplex                 & $92.45_{d=5}$     & $92.23_{d=5}$     & $68.26_{d=50}$    & $69.07_{d=50}$    & $99.53_{d=5}$     & $88.26_{d=24}$    & $94.53_{d=5}$ \\
  Hadamard fixed classifier \cite{hoffer2018fix}      & $73.58_{d=5}$     & $73.38_{d=5}$     & $43.51_{d=50}$    & $43.75_{d=50}$    & $79.67_{d=5}$     & $60.70_{d=24}$    & $74.18_{d=5}$ \\ \hline
  Baseline (learned classifier)                & $92.38_{d=5}$     & $92.19_{d=5}$     & $68.73_{d=50}$    & $68.51_{d=50}$    & $99.22_{d=5}$     & $87.50_{d=24}$    & $94.39_{d=5}$ \\

  \hline\hline
  RePoNets $d$-Simplex                   & $92.53_{d=9}$     & $92.29_{d=9}$     & $68.68_{d=99}$    & $68.31_{d=99}$    & $99.65_{d=9}$     & $88.41_{d=46}$    & $94.51_{d=9}$ \\
  Hadamard fixed classifier \cite{hoffer2018fix}      & $92.03_{d=9}$     & $92.18_{d=9}$     & $67.19_{d=99}$    & $67.85_{d=99}$    & $99.51_{d=9}$     & $88.21_{d=46}$    & $94.53_{d=9}$ \\ \hline
  Baseline (learned classifier)                & $92.19_{d=9}$     & $92.60_{d=9}$     & $68.86_{d=99}$    & $68.46_{d=99}$    & $99.52_{d=9}$     & $88.43_{d=46}$    & $94.26_{d=9}$ \\

  \hline\hline\hline
  Hadamard fixed classifier \cite{hoffer2018fix}      & $90.51_{d=512}$   & $88.28_{d=512}$   & $63.27_{d=512}$   & $64.76_{d=512}$   & $99.50_{d=512}$   & $88.05_{d=512}$   & $94.44_{d=512}$ \\ \hline
  Baseline (learned classifier)                & $92.45_{d=512}$   & $92.52_{d=512}$   & $68.33_{d=512}$   & $68.69_{d=512}$   & $99.52_{d=512}$   & $88.77_{d=512}$   & $94.58_{d=512}$ \\

  \hline
\end{tabular}
\caption{ Reported accuracy (\%) of the RePoNets method vs other methods on different combinations of architectures, feature space dimensions and datasets.}
\label{table:variant}
\vspace{-3mm}
\end{table*}

\subsection{Performance Evaluation}
We evaluate the classification performance of RePoNets on the following datasets: MNIST, EMNIST, FashionMNIST, CIFAR-10 and CIFAR-100. 
The proposed method is compared with the fixed classifier method reported in \cite{hoffer2018fix}, here implemented for different architectures and different dimensions of the embedding space. Standard CNN baselines with learnable classifiers are also included. Except for the final fixed classifier all the compared methods have exactly the same architecture and training settings as the one that RePoNet uses. 

We trained the so called LeNet++ architecture \cite{wen2016discriminative} on all the MNIST family datasets. The network is a modification of the LeNets \cite{lecun1998gradient} to a deeper and wider network including parametric rectifier linear units (pReLU) \cite{he2015delving}.
\begin{figure}[b]
\vspace{-0.2cm}
\centering
\includegraphics[width=0.99\columnwidth]{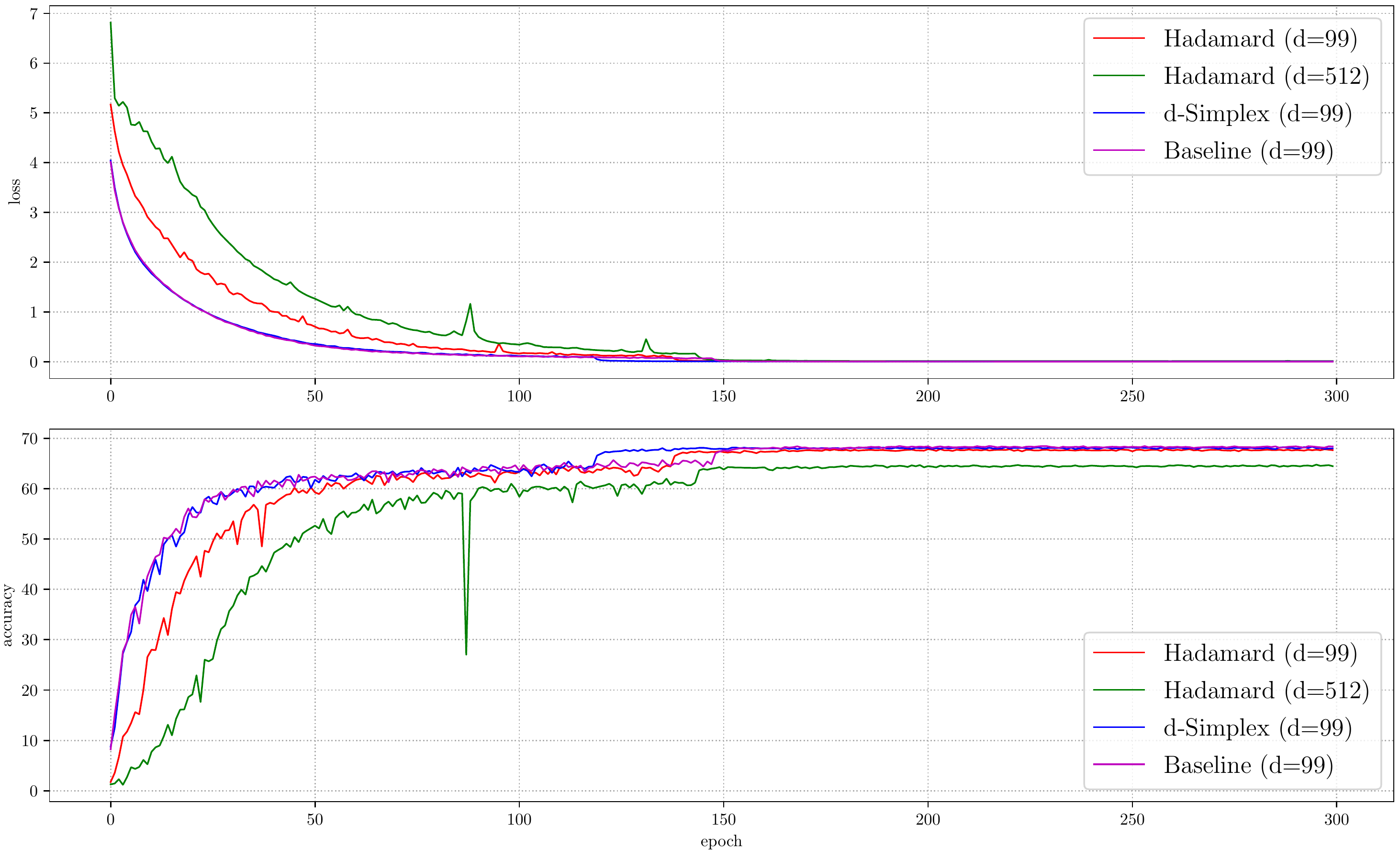}
\caption{Learning speed. Comparing training error (\emph{top}) and test accuracy (\emph{bottom}) on the CIFAR100 dataset. 
\vspace{-0.2cm}
}
\label{fig_learning_speed}
\end{figure}
We further trained the VGG network of \cite{SimonyanZ14a} on the CIFAR-10 and CIFAR-100 datasets using two networks of depth 13 and 19 and the same hyper-parameters used in the original work. Also in this case, we compared all the variants of our approach including the two architectures with different dimensions of the feature space. 

All the results are reported in Tab.~\ref{table:variant}. 
Each entry in the table reports the test-set accuracy. The subscript indicates the specific feature space dimension $d$ used for that experiment. The results reveal and confirm that our fixed classifier models achieve comparable classification accuracy of other trainable classifier models. This evidence is in agreement on all the combinations of datasets, architectures and feature space dimensions. All the RePoNet variants 
exhibit similar behavior even in complex combinations such as in the case of the CIFAR100 dataset in low dimensional feature space. For example, the RePoNet $d$-Cube fixed classifier implemented with the VGG19 architecture achieves an accuracy of $64.02\%$ in a $d=7$ dimensional feature space. A fully trainable classifier in a feature space of dimension $d=512$ (i.e. two orders of magnitude larger) achieves $68.69\%$. With a reasonable shorter feature dimension of $d=50$ RePoNet $d$-Orthoplex improves the accuracy to $69.07\%$.
\setlength{\extrarowheight}{2pt} 
\begin{table}[b]
\vspace{-0.4cm}
	\centering
	\footnotesize 
\begin{tabular}{|c|c|}
  \hline
  
  RePoNets $d$-Cube                 & $66.32_{d=7}$ \\
  Baseline (learned classifier)     & $71.05_{d=7}$ \\

  \hline\hline
  RePoNets $d$-Orthoplex            & $74.32_{d=50}$ \\
  Baseline (learned)                & $74.35_{d=50}$ \\

  \hline\hline
  RePoNets $d$-Simplex              & $73.43_{d=99}$ \\
  Baseline (learned classifier)     & $73.66_{d=99}$ \\

  \hline\hline\hline
  Baseline (learned classifier)     & $73.46_{d=512}$ \\

  \hline
\end{tabular}
\caption{Reported accuracy (\%) of the RePoNets method vs other methods on CIFAR100 with the  DenseNet-169 architecture. 
\vspace{-0.3cm}
}
\label{table:densenet}
\vspace{-3mm}
\end{table}
Results further show that when the number of classes has not an about equal number of unique weight directions in the embedding space (i.e. $d < K$), as in the Hadamard fixed classifier \cite{hoffer2018fix}, no  proper learning can be obtained. This effect is also present for simple datasets as the MNIST digits dataset. 
When $d \approx K$ or $d > K$ classification performance are similar. However, as shown in Fig.~\ref{fig_learning_speed}, RePoNet has faster learning speed matching the trainable baseline. 
Our conjecture is that with our symmetrical fixed classifiers, each term in the loss function tends to have the same magnitude (i.e. von Mises-Fisher distribution is similar to the Normal distribution) and therefore the average is a good estimator. Contrarily, in Hadamard classifier the terms may have different magnitudes and  ``important'' errors may be not taken correctly into account by averaging.  
\\
\indent
We finally evaluate our approach in the more sophisticated DenseNet-169 architecture \cite{Densenet2017}. Tab.\ref{table:densenet} shows that RePoNet classifiers are architecture-agnostic being able to improve accuracy following the expressive power of more competitive architectures.

\section{Conclusion}
We have shown that a special set of fixed classifiers based on regular polytopes generates stationary features by maximally exploiting the available representation space. The proposed method is simple to implement and theoretically correct.
Experimental results confirm both the theoretical analysis and the generalization capability of the approach. RePoNet improves and generalizes the concept of a fixed classifier, recently proposed in \cite{hoffer2018fix}, to a larger class of fixed classifier models exploiting the inherent symmetry of regular polytopes in the feature space.
\\
\indent
Our finding may have implications in all of those Deep Neural Network learning contexts in which a classifier must be robust against changes of the feature representation while learning  
as in incremental and continual learning settings. 


\begin{thebibliography}{10}

\bibitem{hoffer2018fix}
Elad Hoffer, Itay Hubara, and Daniel Soudry.
\newblock Fix your classifier: the marginal value of training the last weight
  layer.
\newblock In {\em International Conference on Learning Representations (ICLR)},
  2018.

\bibitem{Zoph_2018_CVPR}
Barret Zoph, Vijay Vasudevan, Jonathon Shlens, and Quoc~V. Le.
\newblock Learning transferable architectures for scalable image recognition.
\newblock In {\em The IEEE Conference on Computer Vision and Pattern
  Recognition (CVPR)}, June 2018.

\bibitem{Cao18}
Q.~Cao, L.~Shen, W.~Xie, O.~M. Parkhi, and A.~Zisserman.
\newblock Vggface2: A dataset for recognising faces across pose and age.
\newblock In {\em International Conference on Automatic Face and Gesture
  Recognition}, 2018.

\bibitem{goodfellow2016deep}
Ian Goodfellow, Yoshua Bengio, Aaron Courville, and Yoshua Bengio.
\newblock {\em Deep learning}, volume~1.
\newblock MIT Press, 2016.

\bibitem{ranjan2017l2}
Rajeev Ranjan, Carlos~D Castillo, and Rama Chellappa.
\newblock L2-constrained softmax loss for discriminative face verification.
\newblock {\em arXiv preprint arXiv:1703.09507}, 2017.

\bibitem{Liu2017CVPR}
Weiyang Liu, Yandong Wen, Zhiding Yu, Ming Li, Bhiksha Raj, and Le~Song.
\newblock Sphereface: Deep hypersphere embedding for face recognition.
\newblock In {\em CVPR}, 2017.

\bibitem{Liu_2018_CVPR}
Weiyang Liu, Zhen Liu, Zhiding Yu, Bo~Dai, Rongmei Lin, Yisen Wang, James~M.
  Rehg, and Le~Song.
\newblock Decoupled networks.
\newblock In {\em The IEEE Conference on Computer Vision and Pattern
  Recognition (CVPR)}, 2018.

\bibitem{wang2017normface}
Feng Wang, Xiang Xiang, Jian Cheng, and Alan~Loddon Yuille.
\newblock Normface: l 2 hypersphere embedding for face verification.
\newblock In {\em Proceedings of the 2017 ACM on Multimedia Conference}, pages
  1041--1049. ACM, 2017.

\bibitem{liu_2017_coco_v1}
Yu~Liu, Hongyang Li, and Xiaogang Wang.
\newblock Learning deep features via congenerous cosine loss for person
  recognition.
\newblock {\em arXiv preprint: 1702.06890}, 2017.

\bibitem{hasnat2017mises}
Md~Hasnat, Julien Bohn{\'e}, Jonathan Milgram, St{\'e}phane Gentric, Liming
  Chen, et~al.
\newblock von mises-fisher mixture model-based deep learning: Application to
  face verification.
\newblock {\em arXiv preprint arXiv:1706.04264}, 2017.

\bibitem{yuan2017feature}
Yuhui Yuan, Kuiyuan Yang, and Chao Zhang.
\newblock Feature incay for representation regularization.
\newblock {\em arXiv preprint arXiv:1705.10284}, 2017.

\bibitem{wang2018additive}
Feng Wang, Jian Cheng, Weiyang Liu, and Haijun Liu.
\newblock Additive margin softmax for face verification.
\newblock {\em IEEE Signal Processing Letters}, 25(7):926--930, 2018.

\bibitem{Imprinted_Qi_2018_CVPR}
Hang Qi, Matthew Brown, and David~G. Lowe.
\newblock Low-shot learning with imprinted weights.
\newblock In {\em The IEEE Conference on Computer Vision and Pattern
  Recognition (CVPR)}, June 2018.

\bibitem{Wu_2018_ECCV}
Zhirong Wu, Alexei~A. Efros, and Stella~X. Yu.
\newblock Improving generalization via scalable neighborhood component
  analysis.
\newblock In {\em The European Conference on Computer Vision (ECCV)}, September
  2018.

\bibitem{HypersphericalEnergy2018}
Weiyang Liu, Rongmei Lin, Zhen Liu, Lixin Liu, Zhiding Yu, Bo~Dai, and Le~Song.
\newblock Learning towards minimum hyperspherical energy.
\newblock {\em NIPS}, 2018.

\bibitem{thomson1904xxiv}
Joseph~John Thomson.
\newblock {XXIV}. {O}n the structure of the atom: an investigation of the
  stability and periods of oscillation of a number of corpuscles arranged at
  equal intervals around the circumference of a circle; with application of the
  results to the theory of atomic structure.
\newblock {\em The London, Edinburgh, and Dublin Philosophical Magazine and
  Journal of Science}, 7(39):237--265, 1904.

\bibitem{batle2016generalized}
J~Batle, Armen Bagdasaryan, M~Abdel-Aty, and S~Abdalla.
\newblock Generalized thomson problem in arbitrary dimensions and non-euclidean
  geometries.
\newblock {\em Physica A: Statistical Mechanics and its Applications},
  451:237--250, 2016.

\bibitem{tammes1930origin}
Pieter Merkus~Lambertus Tammes.
\newblock On the origin of number and arrangement of the places of exit on the
  surface of pollen-grains.
\newblock {\em Recueil des travaux botaniques n{\'e}erlandais}, 27(1):1--84,
  1930.

\bibitem{bagchi1997stay}
Bhaskar Bagchi.
\newblock How to stay away from each other in a spherical universe.
\newblock {\em Resonance}, 2(9):18--26, 1997.

\bibitem{coxeter1963regular}
H.S.M. Coxeter.
\newblock {\em Regular Polytopes}.
\newblock Macmillan mathematics paperbacks. Macmillan, 1963.

\bibitem{aldous1985exchangeability}
David~J Aldous.
\newblock Exchangeability and related topics.
\newblock In {\em {\'E}cole d'{\'E}t{\'e} de Probabilit{\'e}s de Saint-Flour
  XIII—1983}, pages 1--198. Springer, 1985.

\bibitem{Bernardo1996}
J.M. Bernardo.
\newblock {The concept of exchangeability and its applications}.
\newblock {\em Far East Journal of Mathematical Sciences}, 4:111--122, 1996.

\bibitem{FashionMNIST2017}
Han Xiao, Kashif Rasul, and Roland Vollgraf.
\newblock Fashion-mnist: a novel image dataset for benchmarking machine
  learning algorithms.
\newblock 2017.

\bibitem{EMNIST17}
Gregory Cohen, Saeed Afshar, Jonathan Tapson, and Andr{\'{e}} van Schaik.
\newblock {EMNIST:} extending {MNIST} to handwritten letters.
\newblock In {\em 2017 International Joint Conference on Neural Networks,
  {IJCNN} 2017, Anchorage, AK, USA, May 14-19, 2017}, pages 2921--2926, 2017.

\bibitem{cifar-10}
Alex Krizhevsky.
\newblock Learning multiple layers of features from tiny images, 2009.

\bibitem{kingma2014adam}
Diederik~P. Kingma and Jimmy Ba.
\newblock Adam: {A} method for stochastic optimization.
\newblock In {\em International Conference on Learning Representations (ICLR)},
  2015.

\bibitem{KaimingHe15}
Kaiming He, Xiangyu Zhang, Shaoqing Ren, and Jian Sun.
\newblock Delving deep into rectifiers: Surpassing human-level performance on
  imagenet classification.
\newblock In {\em 2015 {IEEE} International Conference on Computer Vision,
  {ICCV} 2015, Santiago, Chile, December 7-13, 2015}, pages 1026--1034, 2015.

\bibitem{Ioffe17}
Sergey Ioffe and Christian Szegedy.
\newblock Batch normalization: Accelerating deep network training by reducing
  internal covariate shift.
\newblock In Francis Bach and David Blei, editors, {\em Proceedings of the 32nd
  International Conference on Machine Learning}, volume~37 of {\em Proceedings
  of Machine Learning Research}, pages 448--456, Lille, France, 07--09 Jul
  2015. PMLR.

\bibitem{wen2016discriminative}
Yandong Wen, Kaipeng Zhang, Zhifeng Li, and Yu~Qiao.
\newblock A discriminative feature learning approach for deep face recognition.
\newblock In {\em European Conference on Computer Vision}, pages 499--515.
  Springer, 2016.

\bibitem{lecun1998gradient}
Yann LeCun, L{\'e}on Bottou, Yoshua Bengio, and Patrick Haffner.
\newblock Gradient-based learning applied to document recognition.
\newblock {\em Proceedings of the IEEE}, 86(11):2278--2324, 1998.

\bibitem{he2015delving}
Kaiming He, Xiangyu Zhang, Shaoqing Ren, and Jian Sun.
\newblock Delving deep into rectifiers: Surpassing human-level performance on
  imagenet classification.
\newblock In {\em Proceedings of the IEEE international conference on computer
  vision}, pages 1026--1034, 2015.

\bibitem{SimonyanZ14a}
Karen Simonyan and Andrew Zisserman.
\newblock Very deep convolutional networks for large-scale image recognition.
\newblock {\em ICLR}, 2015.

\bibitem{Densenet2017}
G.~Huang, Z.~Liu, L.~v.~d. Maaten, and K.~Q. Weinberger.
\newblock Densely connected convolutional networks.
\newblock In {\em 2017 IEEE Conference on Computer Vision and Pattern
  Recognition (CVPR)}, pages 2261--2269, July 2017.

\end{thebibliography}

\end{document}